\documentclass[sigconf]{aamas}

\setcopyright{ifaamas}  
\acmDOI{}  
\acmISBN{}  
\acmConference[AAMAS'18]{Proc.\@ of the 17th International Conference on Autonomous Agents and Multiagent Systems (AAMAS 2018)}{July 10--15, 2018}{Stockholm, Sweden}{M.~Dastani, G.~Sukthankar, E.~Andr\'{e}, S.~Koenig (eds.)}  
\acmYear{2018}  
\copyrightyear{2018}  
\acmPrice{}  

\usepackage{bm}
\usepackage{url}
\usepackage{amsthm}
\usepackage{amsmath}
\usepackage{amssymb}
\usepackage{appendix}
\usepackage{booktabs}
\usepackage{csquotes}
\usepackage{enumitem}
\usepackage{etoolbox}
\usepackage{multirow}
\usepackage{subcaption}

\renewcommand{\arraystretch}{1.1}

\usepackage{siunitx}
\robustify\bfseries

\setlist[description]{font=\normalfont\itshape\space}

\usepackage{tikz}

\usetikzlibrary{fit}
\usetikzlibrary{calc}
\usetikzlibrary{arrows}

\usepackage{tikz}
\usetikzlibrary{matrix}
\usetikzlibrary{shapes.callouts}
\usepackage{colortbl}

\newcommand*\head[1]{\textbf{#1}}

\definecolor{yellow}{rgb}{0.9,0.95,0}
\definecolor{green}{rgb}{0.2,0.8,0}

\tikzstyle{lobmatrix}=[matrix of nodes,%
    draw=white,%
    nodes={anchor=north,draw=black,inner sep=0pt,minimum height=0.5cm},%
    column sep=1ex,row sep=0.6ex,inner sep=2ex,%
    rounded corners,%
    column 1/.style={minimum width=1.5cm, nodes={}},%
    column 2/.style={minimum width=2.4cm},%
    column 3/.style={minimum width=1.5cm, nodes={}},%
    column 4/.style={minimum width=2.4cm, nodes={}},%
    ampersand replacement=\&
]

\newcommand{\ticker}[1]{\texttt{#1}}

\definecolor[named]{AskRed}{RGB}{224,75,58}
\definecolor[named]{BidBlue}{RGB}{57,139,187}

\DeclareMathOperator{\defeq}{\triangleq}

\DeclareMathOperator{\ControlParam}{\theta}

\DeclareMathOperator{\OrderSize}{\mathrm{Size}}

\DeclareMathOperator{\ReferencePrice}{\mathrm{Ref}}
\DeclareMathOperator{\PriceOffset}{\mathrm{Dist}}
\DeclareMathOperator{\SpreadFactor}{\mathrm{Spread}}

\DeclareMathOperator{\Inventory}{\mathrm{Inv}}

\usepackage{todonotes}

\begin{document}

\title{Market Making via Reinforcement Learning}


\author{Thomas Spooner}
\affiliation{%
    \department{Department of Computer Science}
    \institution{University of Liverpool}
}
\email{t.spooner@liverpool.ac.uk}
\author{John Fearnley}
\affiliation{%
    \department{Department of Computer Science}
    \institution{University of Liverpool}
}
\email{fearnley@liverpool.ac.uk}
\author{Rahul Savani}
\affiliation{%
    \department{Department of Computer Science}
    \institution{University of Liverpool}
}
\email{rahul.savani@liverpool.ac.uk}
\author{Andreas Koukorinis}
\affiliation{Stratagem Technologies Ltd,}
\affiliation{and University College London}
\email{andreas@stratagem.co}

\begin{abstract}
    Market making is a fundamental trading problem in which an agent provides
    liquidity by continually offering to buy and sell a security. The problem is
    challenging due to inventory risk, the risk of accumulating an unfavourable
    position and ultimately losing money. In this paper, we develop a
    high-fidelity simulation of limit order book markets, and use it to design a
    market making agent using temporal-difference reinforcement learning. We use
    a linear combination of tile codings as a value function approximator, and
    design a custom reward function that controls inventory risk. We demonstrate
    the effectiveness of our approach by showing that our agent outperforms both
    simple benchmark strategies and a recent online learning approach from the
    literature.
\end{abstract}

\keywords{Market Making; Limit Order Books; TD Learning; Tile Coding}

\maketitle

%
\section{Introduction}
The role of a market maker is to provide liquidity by
facilitating transactions with other market participants. Like many trading problems, it has
become increasingly automated since the advent of the electronic limit order
book (LOB), as the need to handle more data and act on ever shorter time scales
renders the task almost impossible for humans~\cite{Hasbrouck2013,Leaver2016}.
Upwards of 60\% of trading volume on some particularly active markets has 
been attributed to automated trading
systems~\cite{Savani2012,Haynes2015}. This paper uses reinforcement
learning (RL) to design competitive market making agents for financial markets
using high-frequency historical equities data.

\subsection{Related work.}
Market making has been studied across a number of disciplines, including
economics, finance, artificial intelligence (AI), and machine learning. A
classic approach in the finance literature is to treat market making as a
problem of \emph{stochastic optimal control}. Here, a model for order arrivals
and executions is developed and then control algorithms for the resulting
dynamics are
designed~\cite{Ho1981,Grossman1988,Avellaneda2008,Guilbaud2011,Chakraborty2011,CarteaBLSH}.
Recent results in this line of research have studied price impact, adverse
selection and predictability~\cite{abergel2016limit}, and augmented the problem
characteristics with risk measures and inventory
constraints~\cite{Gueant2013,Cartea2015}.


Another prominent approach to studying market making and limit order book
markets has been that of \emph{zero-intelligence (ZI) agents.} The study of ZI
agents has spanned economics, finance and AI. These agents do not ``observe,
remember, or learn'', but can, for example, adhere to inventory
constraints~\cite{Gode1993}. Newer, more intelligent variants, now even
incorporate learning mechanisms~\cite{Cliff2006,Vytelingum2008}. Here, agents
are typically evaluated in simulated markets without using real market data.

A significant body of literature, in particular in AI, has studied the 
market making problem for \emph{prediction markets}~\cite{Othman2012,Brahma2012,Othman2013}. 
In this setting, the agent's main goal is to elicit information from informed
participants in the market. While later studies have addressed profitability,
the problem setup remains quite distinct from the financial one considered here.


Reinforcement learning has been applied for other financial trading
problems~\cite{Moody1999,Sherstov2004,Schvartzman2009}, including optimal
execution~\cite{Nevmyvaka2006} and foreign exchange trading~\cite{Dempster2006}.
The first case of applying RL to market making~\cite{Chan2001} focused on the
impact of noise (due to uninformed traders) on the agent's quoting behaviour and
showed that RL successfully converges on the expected strategies for a number of
controlled environments. They did not, however, capture the challenges
associated with explicitly handling order placement and cancellation, nor the
complexities of using continuous state variables. Moreover,~\cite{Chan2001}
found that temporal-difference RL struggled in their setting, a finding echoed
in~\cite{Sherstov2004}.~\cite{Chan2001} attributed this to partial observability
and excessive noise in the problem domain, despite the relative simplicity of
their market simulation. In follow up work,~\cite{Shelton2001} used importance
sampling as a solution to the problems observed with off-policy learning. In
contrast, we find temporal-difference RL to be effective for the market making
problem, provided that we use eligibility traces and carefully design our
function approximator and reward function.


One of the most recent related works is~\cite{Abernethy2013}, which uses an
\emph{online learning} approach to develop a market making agent. They prove
nice theoretical results for a stylized model, and empirically evaluate their
agents under strong assumptions on executions. For example, they assume that the
market has sufficient liquidity to execute market orders entirely at the posted
price with no slippage. We use this approach as one of the benchmarks for our
empirical evaluation and address the impact of trading in a more realistic
environment.

\subsection{Our contributions.}
The main contribution of this paper is to design and analyse
temporal-difference (TD) reinforcement learning agents for market making. In
contrast to past work~\cite{Chan2001,Shelton2001} we develop a high-fidelity
simulation using high-frequency historical data. We identify eligibility traces
as a solution to the unresolved issues previously associated with reward
attribution, noise and partial observability~\cite{Chan2001}. We then design a
reward function and state representation and demonstrate that these are key
factors in the success of our final agent. We outline the steps taken to
develop our agent below:

\begin{enumerate}[leftmargin=5mm]
    \item We build a \textbf{realistic, data-driven simulation of a limit order
        book} using a basket of 10 equities across 5 venues and a mixture of
        sectors. This data includes 5 levels of order book depth and transaction
        updates with time increments on the order of milliseconds.
    \item We address concerns raised in past work about the efficacy of one-step
        temporal-difference learning, corroborating their results but
        demonstrating that \textbf{eligibility traces} are a simple and
        effective solution.
    \item We investigate the performance of a wide range of new and old TD-based
        learning algorithms.
    \item We show that the ``natural'' choice of reward function (incremental
        profit and loss) does not lead to the best performance and regularly
        induces instability during learning. We propose a solution in the form
        of an \textbf{asymmetrically dampened} reward function which improves
        learning stability, and produces higher and more consistent returns.
    \item We provide an evaluation of three different state space constructions
        and propose a \textbf{linear combination of tile codings} as our final
        representation as it gives competitive performance with more stable
        learning.
    \item We present a \textbf{consolidated agent}, based on a combination of
        the best results from 1--5 above, and show that it produces the best
        risk-adjusted out-of-sample performance compared to a set of simple
        benchmarks, a basic RL agent, and a recent online learning
        approach~\cite{Abernethy2013}. Moreover, we show that the performance of
        our consolidated agent is competitive enough to represent a potentially
        viable approach for use in practice.
\end{enumerate}

\section{Preliminaries}
\subsection{Limit order books.}
A \emph{limit order} (LO) is an offer to buy or sell a given amount of an asset
at a fixed price (or better). Each limit order specifies a \emph{direction}
(buy/sell or, equivalently, bid/ask), a \emph{price} and a \emph{volume} (how
much to be traded). A \emph{limit order book} is an aggregation of LOs that
have been submitted to the market. The book has a fixed number of \emph{price
levels}, and the gap between the price levels is called the \emph{tick size}.
An example limit order book is shown in Fig.~\ref{fig:lob}.

\begin{figure}
    \centering
    \resizebox{0.7\linewidth}{!}{%
        \begin{tikzpicture}
        \matrix[lobmatrix] (m)
        {
        \& \head{Price} \& \head{Asks}  \\
                    \& 101.00    \& 12   \\
                    \& 100.50    \& 13   \\
                    \& 100.25    \&      \\
        35             \& 100.00     \&      \\
        3              \& 99.75     \&      \\
        11             \& 99.50     \&      \\
        \head{Bids}     \&           \&      \\
        };
        \end{tikzpicture}%
    }

    \caption{Snapshot of a limit order book with multiple price levels occupied
    by bid or ask orders and a total volume.}\label{fig:lob}
\end{figure}

There are two types of orders: \emph{Market Orders} (MOs) and \emph{Limit
Orders}. A market order is a request to buy or sell \emph{now} at the current
market price. When it arrives at the exchange it is executed against limit
orders starting with the best price. That is, a sell market order will execute
against the highest priced bid limit orders, and a buy market order will
execute against the lowest priced ask limit orders. This process may spill-over
to further price levels if there is not enough volume at the first level of the
book.

In contrast, a limit order is an offer to buy or sell at prices no worse than a
limit price. When a limit order arrives at the market, it joins the back of the
execution queue at its quoted price level, meaning that market orders will
match against limit orders at a given price on a first-come first-served basis.
If a limit order is submitted at a price level that currently holds limit
orders with opposite direction (for example, a bid limit order submitted to a
price level with ask limit orders quoted), then they are immediately executed
against each other. This means that a given price level can only have one
direction quoted at any given time.



The \emph{market spread} is the difference between the best bid and ask quotes
in the LOB.\ The \emph{mid-price}, $m$, is the mid-point of the spread, defined
simply by the arithmetic mean of the best bid and ask. When the best bid or ask
changes, the \emph{mid-price move}, denoted by $\Delta m$, is the change in the
mid-price since the last time period.  For example, in Fig.~\ref{fig:lob}, the
best bid and ask prices are 100.00 and 100.50, respectively, the mid-price is
100.25 and the market spread is 0.5. For a full examination of LOBs, their
mechanics and nomenclature, see \citeauthor{Gould2013}~\citeyear{Gould2013} or
\citeauthor{abergel2016limit}~\citeyear{abergel2016limit}.

\subsection{Market making.}
Market makers (MMs) are \textcquote{cartea2015algorithmic}{traders who profit
from facilitating exchange in a particular asset and exploit their skills in
executing trades}. They continually quote prices at which they are willing to
trade on \emph{both} sides of the book and, in so doing, provide liquidity to
the market. If both orders execute, then they gain their \emph{quoted spread},
which will be at least as big as the market spread, but may be more if they did
not quote at the best price on either side.

\emph{Example.} Consider a trader acting in the order book shown in
Fig.~\ref{fig:lob}, with ask and bid orders at 101.00 and 99.50, respectively,
each for $q=100$ units of volume. In this case, the quoted spread is $\Delta =
1.50$ and if both orders execute then the trader will make $q\Delta = 100
\times 1.50 = 150$ profit.

By quoting on both sides of the book, there is a risk that only one of the
orders will execute, and thus the market maker may pick up a non-zero
\emph{inventory}. This may be positive or negative, and represents the extent
to which the market maker has bought more than it has sold, or vice versa. A
simple idealised market maker seeks to make money by repeatedly making its
quoted spread, while keeping its inventory low to minimise market exposure.  A
more advanced market maker may choose to hold a non-zero inventory to exploit
clear trends, whilst simultaneously capturing the spread.



%
\section{The simulator}
We have developed a simulator of a financial market via direct reconstruction
of the limit order book from historical data. This data comprised 10 securities
from 4 different sectors, each with 8 months (January --- August 2010) of data
about the order-book depth, which records how much volume was at each level of
the LOB, and the actual transactions that took place. The simulator tracks the
top 5 price levels in the limit order book, and allows an agent to place its
own orders within these price levels.



This approach to simulating an exchange is highly realistic and makes limited
assumptions about the market. It should be noted that, since the market is
reconstructed from historical data, simulated orders placed by an agent cannot
impact the market. However, for all of the agents studied in this paper, the
size of the orders used by the agent are small compared to the total amount
traded in the market, so the impact of the agent's orders would be negligible.


One problem does arise, however, due the fact that we only have aggregate
information about the limit orders. When we see that the amount of volume at a
particular price level has decreased, we know that either some of the orders at
that level have either been executed or they have been cancelled. Since we have
transaction data, we can deal with the case where limit orders have been
executed, but when limit orders are cancelled, we do not know precisely
\emph{which} orders have been removed.

This causes a problem when an agent's order is currently being simulated for
that price level, because we do not know whether the cancelled order was ahead
or behind the simulated order in the queue. Our solution is to assume that
cancellations are distributed uniformly throughout the queue. This means that
the probability that the cancelled order is ahead of the agent's order is
proportional to the amount of volume ahead of the agent's order compared to the
amount of volume behind it.

\section{The agent}
\subsection{Trading strategy.}
In this paper we consider a state-based market making agent that acts on events
as they occur in the LOB, subject to constraints such as upper and lower limits
on inventory. An event may occur due to anything from a change in price, volume
or arrangement of orders in the book; anything that constitutes an observable
change in the state of the environment. Importantly, this means that the
agent's actions are not spaced regularly in time. Since we are building a
market maker, the agent is required to quote prices at which it is willing to
buy and sell at all valid time points, unless the inventory constraints are no
longer satisfied at which point trading is restricted to orders that bring the
agent closer to a neutral position.

\begin{table}
    \centering
    \caption{A typical action-space for market making.}\label{tab:action_space}

    \begin{tabular}{lccccc|cc|cc}
        \toprule\toprule
        Action ID & 0 & 1 & 2 & 3 & 4 & 5 & 6 & 7 & 8 \\
        \midrule
        Ask ($\ControlParam_\text{a}$) & 1 & 2 & 3 & 4 & 5 & 1 & 3 & 2 & 5 \\
        Bid ($\ControlParam_\text{b}$) & 1 & 2 & 3 & 4 & 5 & 3 & 1 & 5 & 2 \\
        \midrule
        Action 9 & \multicolumn{8}{l}{MO with $\OrderSize_\text{m} =
        -\Inventory(t_i)$} \\
        \bottomrule\bottomrule
    \end{tabular}
\end{table}

The action-space (Table~\ref{tab:action_space}) is based on a typical market
making strategy in which the agent is restricted to a single buy and sell order
and cannot exit the market, much like that considered by
\citeauthor{Chakraborty2011}~\citeyear{Chakraborty2011}. There are ten
available actions, the first nine (IDs 0--8) of which correspond to a pair of
orders with a particular spread and bias in their prices. Smaller values of
$\theta_\text{a,b}$ lead to quotes that are closer to the top of the book,
while larger numbers cause the quotes to be deeper in the book. This means that
the agent may choose to \emph{quote wide or tight}, or even to \emph{skew its
orders} in favour of the buy/sell side. The final action (ID 9) allows the
agent to \emph{clear its inventory} using a market order.

Formally, actions~0 through~8 cause limit orders to be placed at fixed
distances relative to a reference price, $\ReferencePrice(t_i)$, chosen here to
be a measure of the true market price: the mid-price.  At each time step,
$t_i$, the agent revises two control parameters, $\ControlParam_\text{a}(t_i)$
and $\ControlParam_\text{b}(t_i)$, from which the relative prices,
$\PriceOffset_\text{a,b}(t_i)$, are derived; the allowed values are given in
Table~\ref{tab:action_space}. The full specification of the agent's pricing
strategy is thus given by Eqs.~\ref{eq:model}~and~\ref{eq:model:sb} below,
\begin{align}
    p_\text{a,b}(t_i) &= \ReferencePrice(t_i) + \PriceOffset_\text{a,b}(t_i),
    \label{eq:model} \\
    \PriceOffset_\text{a,b}(t_i) &= \ControlParam_\text{a,b}(t_i) \cdot
    \SpreadFactor(t_i), \label{eq:model:sb}
\end{align}
where $\SpreadFactor(t_i)$ is a time-dependent scale factor which must be a
multiple of the market's tick size. In the experiments to follow,
$\SpreadFactor(t_i)$ is calculated by taking a moving average of the market
half-spread: $s(t_i)/2$.

All limit orders are placed with fixed, non-zero integer volumes,
$\OrderSize_\text{a,b}$.
The market order used in action~9 is sized proportionately to the agent's current inventory,
$\OrderSize_\text{m}(t_i) = -\alpha\cdot\Inventory(t_i)$, where $\alpha \in (0,
1]$ is a scaling factor and $\Inventory(t_i)$ is the agent's inventory at time
$t_i$. We ensure that $\OrderSize_\text{m}$ is restricted to be a non-zero
integer multiple of the minimum lot size in the market. Note that volume is
defined to be negative for buy orders and positive for sell orders, as in the
nomenclature of \citeauthor{Gould2013}~\citeyear{Gould2013}.

\subsection{Reward functions.}
The ``natural'' reward function for trading agents is profit and loss (PnL):
i.e.\ how much money is made or lost through exchanges with the market. Under
this reward function, the agent is encouraged to execute trades and hold an
inventory which will appreciate in value over time; this is captured in
Eq.~\ref{eq:incremental_pnl}.

Formally, for a given time $t_i$, let $\text{Matched}_\text{a}(t_i)$ and
$\text{Matched}_\text{b}(t_i)$ be the amount of volume \emph{matched}
(\emph{executed}) against the agent's orders since the last time $t_{i-1}$ in
the ask and bid books, respectively, and let $m(t_i)$ denote the mid-price at
time $t_i$. We define
\begin{align*}
    \psi_\text{a}(t_i) &\defeq \text{Matched}_\text{a}(t_i) \cdot
    \left[p_\text{a}(t_i) - m(t_i)\right], \\
    \psi_\text{b}(t_i) &\defeq \text{Matched}_\text{b}(t_i) \cdot \left[m(t_i)
    - p_\text{b}(t_i)\right],
\end{align*}
where $p_\text{a,b}(t_i)$ are given by Eq.~\ref{eq:model}. These functions
compute the money lost or gained through executions of the agent's orders
\emph{relative} to the mid-price. Observe that if both orders are executed in
the same time interval $[t_{i-1}, t_i]$, then $\psi_\text{a}(t_i) +
\psi_\text{b}(t_i)$ simplifies to the agent's quoted spread.  Then, if
$\Inventory(t_{i})$ denotes the agent's inventory at time $t_i$, we can define
the incremental (non-dampened) PnL function $\Psi(t_i)$ by setting $\Psi(t_0)
\defeq 0$ and
\begin{equation}\label{eq:incremental_pnl}
    \Psi(t_i) \defeq \psi_\text{a}(t_i) + \psi_\text{b}(t_i) +
    \Inventory(t_{i}) \Delta m(t_i).
\end{equation}
The third (inventory) term here corresponds to the change in the agent's cash
holdings due to exposure to changes in price in the market. Note that this term
is only necessary because we accounted for the PnL from the agent's trades
relative to the mid-price.

While the definition given above is a natural choice for the problem domain, it
will be shown in this paper that such a basic formulation of the reward
function ignores the specific objectives of a market maker, often leading to
instability during learning and unsatisfactory out-of-sample performance. We
thus study two alternative, \emph{dampened} definitions of reward which are
engineered to disincentivise trend-following and bolster spread capture. The
three reward functions that we study are as follows:

\begin{description}\itemsep2mm
    \item[\hspace{1em}PnL:] \begin{equation}
        r_i = \Psi(t_i). \label{eq:reward:basic}
    \end{equation}
    \item[\hspace{1em}Symmetrically dampened PnL:] \begin{equation}
        r_i = \Psi(t_i) - \eta\cdot\Inventory(t_{i})\Delta m(t_i).
        \label{eq:reward:symm}
    \end{equation}
    \item[\hspace{1em}Asymmetrically dampened PnL:] \begin{equation}
        r_i = \Psi(t_i) - \max[0,\,\eta\cdot\Inventory(t_{i})\Delta m(t_i)].
        \label{eq:reward:asymm}
    \end{equation}
\end{description}
For the final two reward functions, the dampening is applied to the inventory
term, using a scale factor $\eta$. Intuitively, this reduces the reward that
the agent can gain from speculation. Specifically, the symmetric version
dampens both profits and losses from speculation, while asymmetric dampening
reduces the profit from speculative positions but keeps losses intact. In both
cases, the amount of reward that can be gained from capturing the spread
increases relative to the amount of reward that can be gained through
speculation, thus encouraging market making behaviour.



\subsection{State representation.}
The state of the environment is constructed from a set of attributes that
describe the condition of the agent and market, called the \emph{agent-state}
and \emph{market-state}, respectively. The agent-state is described by three
variables, the first is \textbf{inventory}, $\Inventory(t_i)$ --- the amount of
stock currently owned or owed by the agent; equivalently, one may think of the
agent's inventory as a measure of \emph{exposure}, since a large absolute
position opens the agent to losses if the market moves. Next are the
\textbf{active quoting distances}, normalised by the current spread scale
factor, $\SpreadFactor(t_i)$; these are the effective values of the control
parameters, $\ControlParam_\text{a,b}$, after stepping forward in the
simulation.

The greatest source of state complexity comes from the market. Unlike the
agent's internal features, the market state is subject to partial observability
and may not have a Markovion representation. As such, the choice of variables,
which may be drawn from the extensive literature on market microstructure, must
balance expressivity with informational value to avoid Bellman's curse of
dimensionality. In this paper, we include the following:
\begin{enumerate}
    \item Market (bid-ask) spread ($s$),
    \item Mid-price move ($\Delta m$),
    \item Book/queue imbalance,
    \item Signed volume,
    \item Volatility,
    \item Relative strength index.
\end{enumerate}

Three different representations of state, based on these parameters, are
considered. The first two, named the \emph{agent-state} and \emph{full-state},
are represented using a conventional tile coding approximation scheme. The
third takes a more involved approach, whereby the agent-state, market-state and
full-state are all approximated simultaneously using independent tile codings,
and the value is a (fixed) linear combination of the values in these three tile
codings. We refer to this as a linear combination of tile codings (LCTC).

\emph{Linear combination of tile codings.}
Assume there are $N$ independent tile codings. Let $\hat v_i(x)$ be the value
estimate of the state, $x$, given by the $i$-th tile coding. The value of a
state is then defined by the following sum:
\begin{equation}
    \hat v(x) = \sum_{i=0}^{N-1} \lambda_i \hat v_i(x) = \sum_{i=0}^{N-1}
        \lambda_i \sum_{j=0}^{n_i-1} b_{ij}(x) \, w_{ij},
\end{equation}
where $n_i$ is the number of tiles in the $i$-th tile coding, $w_{ij}$ is the
$j$-th value of the associated weight vector and $b_{ij}$ is the binary
activation of the $j$-th tile (taking a value 0 or 1). We enforce that the
$\lambda_i$ weights (or influences) sum to unity for simplicity:
$\sum_{i=0}^{N-1} \lambda_i = 1$.

This approach amounts to learning three independent value functions, each of
which is updated using the same TD-error for gradient descent. Related
approaches involve using multiple tile codings with varying granularity, but
using the same set of state variables, and specific applications to dimension
mapping for dynamics modelling~\cite{Grzes2010,Duan1999}. It has been argued
that a coarse representation improves learning efficiency of a high resolution
representation by directly the agent towards more optimal regions of policy
space. The justification here is the same, the crucial difference is that we
exploit the (approximate) \emph{independency of variables} governing the
agent-state and market-state rather than the shape of features; though these
techniques could be combined.

This technique is particularly relevant for problems where a lot of
domain-specific knowledge is available a priori. For example, consider a
trading agent holding an inventory of, say, 100 units of a given stock. Though
the expected future value of the holdings are surely conditional on the state
of the market, the most important factor for the agent to consider is the risk
associated with being exposed to unpredictable changes in price. Learning the
value for the agent-, market- and full-state representations independently
enables the agent to learn this much faster since it does not rely on observing
every permutation of the full-state to evaluate the value of it's inventory. It
is plausible that this will also help the agent converge (as in the granularity
version mentioned previously) by guiding the agent away from local optima in
policy space.

\subsection{Learning algorithms.}
We consider Q-learning (QL), SARSA and R-learning~\cite{Sutton1998}. Several
variants on these are also evaluated, including: Double
Q-learning~\cite{Hasselt2010}, Expected SARSA~\cite{Harmvan2009}, On-policy
R-learning and Double R-learning. Though other work has taken the approach of
developing a specialist algorithm for the domain~\cite{Nevmyvaka2006}, these
general purpose algorithms have been shown to be effective across a wide array
of problems. Further, by including discounted and undiscounted learning methods
we may identify characteristics of the problem domain that will guide our
approach.

It should be assumed throughout this paper that each algorithm was implemented
using eligibility traces (ETs), unless otherwise stated. As such, the
conventional suffix, as used by Q$(\lambda)$ versus Q-learning, shall be
omitted in favour of uniformity\footnote{The code is provided open source at
\url{https://github.com/tspooner/rl_markets}.}.

%
\section{Results}
We begin by introducing a set of benchmarks: a group of fixed ``spread-based''
strategies, and an online learning approach based on the work of
\citeauthor{Abernethy2013}~\citeyear{Abernethy2013} to anchor the performance
of our agents in the literature. Next, a pair of \emph{basic agents}, using an
agent-state representation, non-dampened PnL reward function and QL or SARSA,
will be presented. These basic agents represent the best attainable results
using ``standard techniques'' and are closely related to those introduced
by~\citeauthor{Chan2001}~\citeyear{Chan2001}. We then move on to an assessment
of our proposed \emph{extensions}, each of which is evaluated and compared to
the basic implementation. This gives an idea of how each of our modifications
perform separately. Finally, we present a \emph{consolidated agent} combining
the most effective methods and show that this approach produces the best
out-of-sample performance relative to the benchmarks, the basic agent, and in
absolute terms.

\begin{table}
    \centering
    \caption{Default parameters as used by the learning algorithm and the
    underlying trading strategy.}\label{tab:params}

\scalebox{0.9}{%
    \begin{tabular}{@{}ll@{}}
        \toprule\toprule
        & Value \\
        \midrule
        Training episodes & 1000 days \\
        Training sample size & $\sim 120$ days \\
        Testing sample size & 40 days \\
        \midrule
        Memory size & $10^7$ \\
        Number of tilings ($M$) & 32 \\
        Weights for linear combination of tile codings &
        \multirow{2}{*}{$(0.6, 0.1, 0.3)$} \\
        $[\text{agent, market, full}]$ ($\lambda_i$) & \\
        \midrule
        Learning rate ($\alpha$) & 0.001 \\
        Step-size [R-learning] ($\beta$) & 0.005 \\
        Discount factor ($\gamma$) & 0.97 \\
        Trace parameter ($\lambda$) & 0.96 \\
        \midrule
        Exploration rate ($\varepsilon$) & 0.7 \\
        $\varepsilon_\text{Floor}$ & 0.0001 \\
        $\varepsilon_\text{T}$ & 1000 \\
        \midrule
        Order size ($\omega$) & 1000 \\
        Min inventory ($\min \Inventory$) & -10000 \\
        Max inventory ($\max \Inventory$) & 10000 \\
        \bottomrule\bottomrule
    \end{tabular}
}
\end{table}

\emph{Experimental setup.}
In each experiment, the dataset was split into disjoint training and testing
sets, where all of the testing data occurs chronologically later than the
training data; separate validation sets were used for hyperparameter
optimisation and for drawing comparisons between the various agents considered
in this paper. Unless otherwise stated, it should be assumed that each of these
experiments were conducted using the parameters quoted in
Table.~\ref{tab:params} which were chosen through a combination of good
practices and the use of a genetic algorithm (GA). The GA was specifically
applied to find effective combinations of state variables, their lookback
parameters and the weights used with the LCTC.

\emph{Performance criteria.}
The primary metric used to assess the performance of a trading strategy is the
amount of money that it makes. While it is common to do this in terms of
\emph{returns} (final excess capital divided by starting capital), this does
not make sense for a market maker since there is no natural notion of starting
capital. We also cannot rate our strategies based on \emph{profit} since each
agent is tested across a variety of stocks, each of which have varying prices
and liquidity. Instead, we introduce a \emph{normalised daily PnL} that rates
how good our strategies are at capturing the spread. This metric is defined on
a daily basis as the total profit divided by the average market spread which
normalises the profit across different markets; this amounts to the number of
market spreads that would need to be captured to obtain that profit.

Ideally, market makers should not keep large inventories. To measure how well
our agents achieve this, the \emph{mean absolute position} (MAP) held by the
agent is quoted as well. For example, high values for this metric may indicate
that the agent has taken a speculative approach to trading. On the other hand,
small values could suggest that the agent relies less on future changes in
market value to derive it's earnings.

Uncertainties are quoted using the standard deviation and mean absolute
deviation for normalised PnL and MAP, respectively.

\subsection{Benchmarks.}
To compare with existing work, we include a spread-based benchmark strategy
which uses an online learning meta-algorithm; the original version, introduced
by~\citeauthor{Abernethy2013}~\citeyear{Abernethy2013}, was adapted to work in
our simulator. We consider the \emph{MMMW} variant which makes use of the
multiplicative weights algorithm to pick in each period from a class of simple
strategies parametrised by a minimum quoted spread\footnote{Consistent with the
original work, MMMW was found to outperform the follow-the-leader variant, as
shown in Table~\ref{tab:ftl}.}. The results which are shown in
Table~\ref{tab:main_comparison} present less favourable performance than the
original paper which found the strategy to be profitable over all dates and
stocks considered. This could be attributed to the use of a less realistic
market simulation that did not, for example, track the limit order book to the
same level of precision considered here. This may indicate that their results
do not generalise to more realistic markets.

A set of benchmarks were then developed using simple and random policies over
the market making action space in Table~\ref{tab:action_space}. These
strategies quote at fixed, symmetric distances from the chosen reference price
($\ControlParam_\text{a} = \ControlParam_\text{b}$) at all times. Unlike the
previous benchmark strategy, this approach accounts for liquidity in the market
by adapting it's prices to changes in the bid-ask spread. Further, in all of
these strategies, market orders are used to forcibly clear the agent's
inventory if the upper or lower bound is reached.

\begin{table}
    \centering
    \caption{Out-of-sample normalised daily PnL (ND-PnL) and mean absolute
    position (MAP) for fixed and random benchmark strategies with inventory
    clearing, evaluated on \ticker{HSBA.L}.}\label{tab:benchmarks:hsba:clearing}

\scalebox{0.9}{%
    \begin{tabular}{
        l
        S[table-format=-2.2,table-figures-uncertainty=1,
        separate-uncertainty=true,table-align-uncertainty=true]
        S[table-format=3, table-figures-uncertainty=1,
        separate-uncertainty=true,table-align-uncertainty=true]
    }
        \toprule\toprule
        & {ND-PnL [$10^4$]} & {MAP [units]} \\
        \midrule
        \textbf{Fixed} ($\ControlParam_\text{a,b} = 1$) & -20.95 \pm 17.52 &
        3646 \pm 2195 \\
        \textbf{Fixed} ($\ControlParam_\text{a,b} = 2$) & 2.97 \pm 13.12 & 3373
        \pm 2181 \\
        \textbf{Fixed} ($\ControlParam_\text{a,b} = 3$) & 0.42 \pm 9.62 & 2674
        \pm 1862 \\
        \textbf{Fixed} ($\ControlParam_\text{a,b} = 4$) & 1.85 \pm 10.80 & 2580
        \pm 1820 \\
        \textbf{Fixed} ($\ControlParam_\text{a,b} = 5$) & \bfseries 2.80 \pm
        10.30 & 2678 \pm 1981 \\
        \textbf{Random} (Table~\ref{tab:action_space}) & -10.82 \pm 5.63 &
        \bfseries 135 \pm 234 \\
        \bottomrule\bottomrule
    \end{tabular}%
}
\end{table}

A sample of the results is given in Table~\ref{tab:benchmarks:hsba:clearing}
for \ticker{HSBA.L}; agent performance was found to be consistent across the
full dataset. Fixed strategies with $\ControlParam_\text{a,b} > 1$ were found
to be just profitable on average, with decreasing MAP as
$\ControlParam_\text{a,b}$ increases.
In all cases we find that the strategy suffers from high volatility. This may
have been caused by a lack of proper inventory management, as indicated by the
consistently high mean average positions.

\subsection{Basic agent.}
The basic agent is used to bound the expected performance \emph{without} the
extensions proposed in this paper. This agent used a state representation
comprising only of the agent-state --- inventory and normalised bid/ask quoting
distances --- and the non-dampened PnL reward function
(Eq.~\ref{eq:reward:basic}). This agent was trained using one-step Q-learning
and SARSA and, consistent with the findings of past work~\cite{Chan2001}, were
unable to obtain useful results in either case.  The addition of eligibility
traces (known as Q$(\lambda)$ and SARSA$(\lambda)$) improved the agents'
performance, yielding strategies that occasionally generated profits
out-of-sample as seen in Table~\ref{tab:basic:return}. These results, however,
were still unsatisfactory.


\begin{table}
    \centering
    \caption{Mean and standard deviation on the normalised daily PnL (given in
    units of $10^4$) for Q-learning and SARSA using non-dampened PnL reward
    function and agent-state.}\label{tab:basic:return}

\scalebox{0.9}{%
    \begin{tabular}{
        l
        S[table-format=-2.2, table-figures-uncertainty=1,
        separate-uncertainty=true,table-align-uncertainty=true]
        S[table-format=-2.2, table-figures-uncertainty=1,
        separate-uncertainty=true,table-align-uncertainty=true]
    }
        \toprule\toprule
        & {QL} & {SARSA} \\
        \midrule
        \ticker{CRDI.MI} & \bfseries 8.14 \pm 21.75 & 4.25 \pm 42.76 \\
        \ticker{GASI.MI} & -4.06 \pm 48.36 & \bfseries 9.05 \pm 37.81 \\
        \ticker{GSK.L} & 4.00 \pm 89.44 & \bfseries 13.45 \pm 29.91 \\
        \ticker{HSBA.L} & \bfseries -12.65 \pm 124.26 & -12.45 \pm 155.31 \\
        \ticker{ING.AS} & -67.40 \pm 261.91 & \bfseries -11.01 \pm 343.28 \\
        \ticker{LGEN.L} & \bfseries 5.13 \pm 36.38 & 2.53 \pm 37.24 \\
        \ticker{LSE.L} & 4.40 \pm 16.39 & \bfseries 5.94 \pm 18.55 \\
        \ticker{NOK1V.HE} & \bfseries -7.65 \pm 34.70 & -10.08 \pm 52.10 \\
        \ticker{SAN.MC} & -4.98 \pm 144.47 & \bfseries 39.59 \pm 255.68 \\
        \ticker{VOD.L} & \bfseries 15.70 \pm 43.55 & 6.65 \pm 37.26 \\
        \bottomrule\bottomrule
    \end{tabular}%
}
\end{table}


\subsection{Extensions.}
In this section, three extensions to the basic agent are discussed: variations
on the learning algorithm, reward function and state representation. The
results are given in Table~\ref{tab:extensions:return}. Each row of the table
consists of the basic agent, \emph{again only agent variables included in the
state space}, with the one modifier. SARSA with eligibility traces is used as
base the learning algorithm, unless stated otherwise.


\begin{table*}
    \centering
    \caption{Mean and standard deviation on the normalised daily PnL (given in
    units of $10^4$) for various extensions to the basic agent, each evaluated
    over the basket of stocks.}\label{tab:extensions:return}

    \renewcommand{\arraystretch}{1.25}
    \resizebox{\linewidth}{!}{%
    \begin{tabular}{
        l
        S[table-format=-3.2, table-figures-uncertainty=1,
        separate-uncertainty=true,table-align-uncertainty=true]
        S[table-format=-3.2, table-figures-uncertainty=1,
        separate-uncertainty=true,table-align-uncertainty=true]
        S[table-format=-3.2, table-figures-uncertainty=1,
        separate-uncertainty=true,table-align-uncertainty=true]
        S[table-format=-3.2, table-figures-uncertainty=1,
        separate-uncertainty=true,table-align-uncertainty=true]
        S[table-format=-3.2, table-figures-uncertainty=1,
        separate-uncertainty=true,table-align-uncertainty=true]
        S[table-format=-3.2, table-figures-uncertainty=1,
        separate-uncertainty=true,table-align-uncertainty=true]
        S[table-format=-3.2, table-figures-uncertainty=1,
        separate-uncertainty=true,table-align-uncertainty=true]
        S[table-format=-3.2, table-figures-uncertainty=1,
        separate-uncertainty=true,table-align-uncertainty=true]
        S[table-format=-3.2, table-figures-uncertainty=1,
        separate-uncertainty=true,table-align-uncertainty=true]
        S[table-format=-3.2, table-figures-uncertainty=1,
        separate-uncertainty=true,table-align-uncertainty=true]
    }
        \toprule\toprule
        & {\ticker{CRDI.MI}} & {\ticker{GASI.MI}} & {\ticker{GSK.L}} &
        {\ticker{HSBA.L}} & {\ticker{ING.AS}} & {\ticker{LGEN.L}} &
        {\ticker{LSE.L}} & {\ticker{NOK1V.HE}} & {\ticker{SAN.MC}} &
        {\ticker{VOD.L}} \\
        \midrule
        \textbf{Double Q-learning} & -5.04 \pm 83.90 & 5.46 \pm 59.03 & 6.22
        \pm 59.17 & 5.59 \pm 159.38 & 58.75 \pm 394.15 & 2.26 \pm 66.53 & 16.49
        \pm 43.10 & -2.68 \pm 19.35 & 5.65 \pm 259.06 & 7.50 \pm 42.50 \\
        \textbf{Expected SARSA} & 0.09 \pm 0.58 & 3.79 \pm 35.64 & -9.96 \pm
        102.85 & 25.20 \pm 209.33 & 6.07 \pm 432.89 & 2.92 \pm 37.01 & 6.79 \pm
        27.46 & -3.26 \pm 25.60 & 32.28 \pm 272.88 & 15.18 \pm 84.86 \\
        \textbf{R-learning} & 5.48 \pm 25.73 & -3.57 \pm 54.79 & 12.45 \pm
        33.95 & -22.97 \pm 211.88 & -244.20 \pm 306.05 & -3.59 \pm 137.44 &
        8.31 \pm 23.50 & -0.51 \pm 3.22 & 8.31 \pm 273.47 & 32.94 \pm 109.84 \\
        \textbf{Double R-learning} & 19.79 \pm 85.46 & -1.17 \pm 29.49 & 21.07
        \pm 112.17 & -14.80 \pm 108.74 & 5.33 \pm 209.34 & -1.40 \pm 55.59 &
        6.06 \pm 25.19 & 2.70 \pm 15.40 & 32.21 \pm 238.29 & 25.28 \pm 92.46 \\
        \textbf{On-policy R-learning} & 0.00 \pm 0.00 & 4.59 \pm 17.27 & 14.18
        \pm 32.30 & 9.56 \pm 30.40 & 18.91 \pm 84.43 & -1.14 \pm 40.68 & 5.46
        \pm 12.54 & 0.18 \pm 5.52 & 25.14 \pm 143.25 & 16.30 \pm 32.69 \\
        \midrule
        \textbf{Symm. Damp.} ($\eta = 0.6$) & 12.41 \pm 143.46 & 9.07 \pm 68.39
        & 30.04 \pm 135.89 & -11.80 \pm 214.15 & 90.05 \pm 446.09 & 5.54 \pm
        119.86 & 8.62 \pm 27.23 & -4.40 \pm 84.93 & 27.38 \pm 155.93 & 8.87 \pm
        93.14 \\
        \textbf{Asymm. Damp.} ($\eta = 0.6$) & 0.08 \pm 2.21 & -0.10 \pm 1.04 &
        9.59 \pm 10.72 & 13.88 \pm 10.60 & -6.74 \pm 68.80 & 4.08 \pm 7.73 &
        1.23 \pm 1.80 & 0.52 \pm 3.29 & 5.79 \pm 13.24 & 9.63 \pm 6.94 \\
        \midrule
        \textbf{Full-state} & -31.29 \pm 27.97 & -35.83 \pm 13.96 & -31.29 \pm
        27.97 & -84.78 \pm 31.71 & -189.81 \pm 68.31 & -14.39 \pm 9.38 & -6.76
        \pm 11.52 & -9.30 \pm 23.17 & -144.70 \pm 104.64 & -21.76 \pm 17.71 \\
        \textbf{LCTC-state} & -5.32 \pm 52.34 & 5.92 \pm 40.65 & 5.45 \pm
        40.79 & -0.79 \pm 68.59 & 9.00 \pm 159.91 & 6.73 \pm 22.88 & 3.04 \pm
        5.83 & -2.72 \pm 19.23 & 52.55 \pm 81.70 & 7.02 \pm 48.80 \\
        \bottomrule\bottomrule
    \end{tabular}%
    }
    \renewcommand{\arraystretch}{1.1}
\end{table*}


\emph{Learning algorithms.}
The first variation on the basic agent deals with the impact of using different
learning algorithms to QL and SARSA;\ the findings are summarised in
Table~\ref{tab:extensions:return}. It was found that variants based on
off-policy learning tended to perform worse than their on-policy counterparts.
Note, for example, the difference between the consistency and outright
performance between R-learning and on-policy R-learning. This may be attributed
to known divergence issues with classical off-policy methods.

Though variants were found to outperform the basic agent for some specific
equity, none were as consistent as SARSA.\ In the majority of cases they
performed worse, both in out-of-sample performance and learning efficiency. For
example, Double Q-learning, despite mitigating the maximisation
bias~\cite{Hasselt2010} versus regular Q-learning, was found to be less
profitable and more volatile on most assets; save for \ticker{CRDI.MI}, a
highly active and volatile stock which may benefit from the double-estimator
approach. This again suggests that, while each stock could be optimised
individually for maximal performance, SARSA may be used reliably as a baseline.

\begin{figure}
    \centering
    \includegraphics[width=0.9\linewidth]{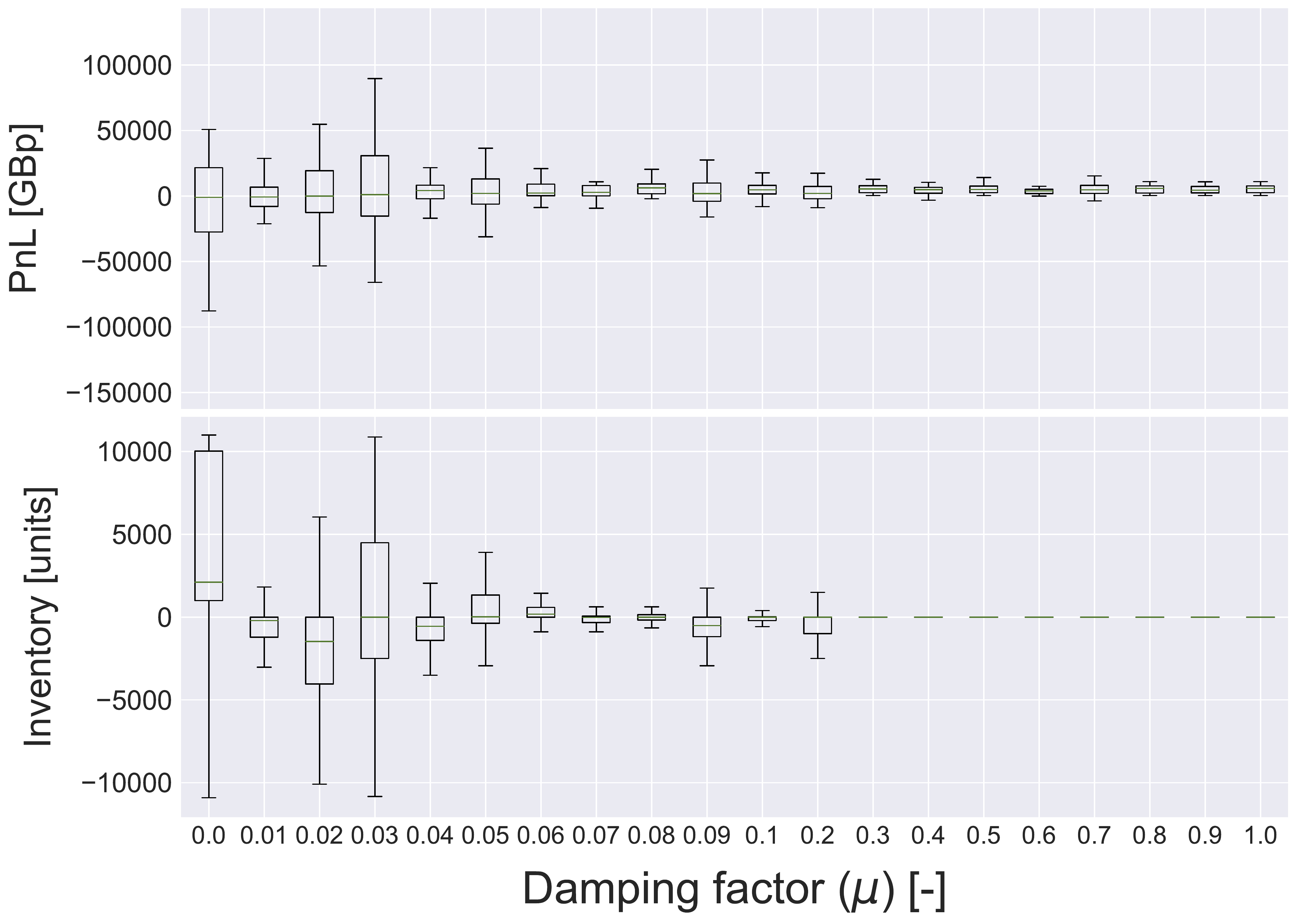}

    \caption{Distributions of daily out-of-sample PnL and mean inventory
    for increasing values of the damping factor, $\eta$, evaluated on
    \ticker{HSBA.L}.}\label{fig:eta_dists}
\end{figure}

\emph{Reward functions.} The results in Table~\ref{tab:extensions:return} show
that the intuitive choice of reward function (non-dampened PnL) does not equate
to the best out-of-sample performance across the basket of securities. Though
symmetric dampening was found to exacerbate the flaws in the basic agent,
asymmetric dampening of the trend-following term in
Eq.~\ref{eq:incremental_pnl}, with sufficiently high damping factor ($\eta$),
was found to produce superior risk-adjusted performance in most cases. This is
exemplified by Fig.~\ref{fig:eta_dists} which shows the distribution of
out-of-sample (actual) PnL for increasing values of the damping factor.

Fig.~\ref{fig:eta_dists} suggests that there is a value of $\eta \sim 0.1$
about which the agent begins to converge on fundamentally different policies
than those promoted by the non-dampened PnL function
(Eq.~\ref{eq:incremental_pnl}). This shift in the final policy is manifested in
the change from holding large, biased inventories towards small and neutral
positions. This reduction in exposure, beginning from $\eta \sim 0.05$, comes
along with a change in PnL, which also tends towards lower variance as a
result. Though the precise value of $\eta$ varies across the basket of
securities, the impact of asymmetric dampening is clear, providing strong
evidence that the inventory term in Eq.~\ref{eq:incremental_pnl} is the leading
driver of agent behaviour and should be manipulated to match one's risk
profile. This result relates strongly with the use of inventory penalising
terms in the value function in stochastic optimal
control~\cite{Cartea2015,Cartea2016} and is a key contribution of this paper.

\begin{figure}
    \centering
    \includegraphics[width=0.9\linewidth]{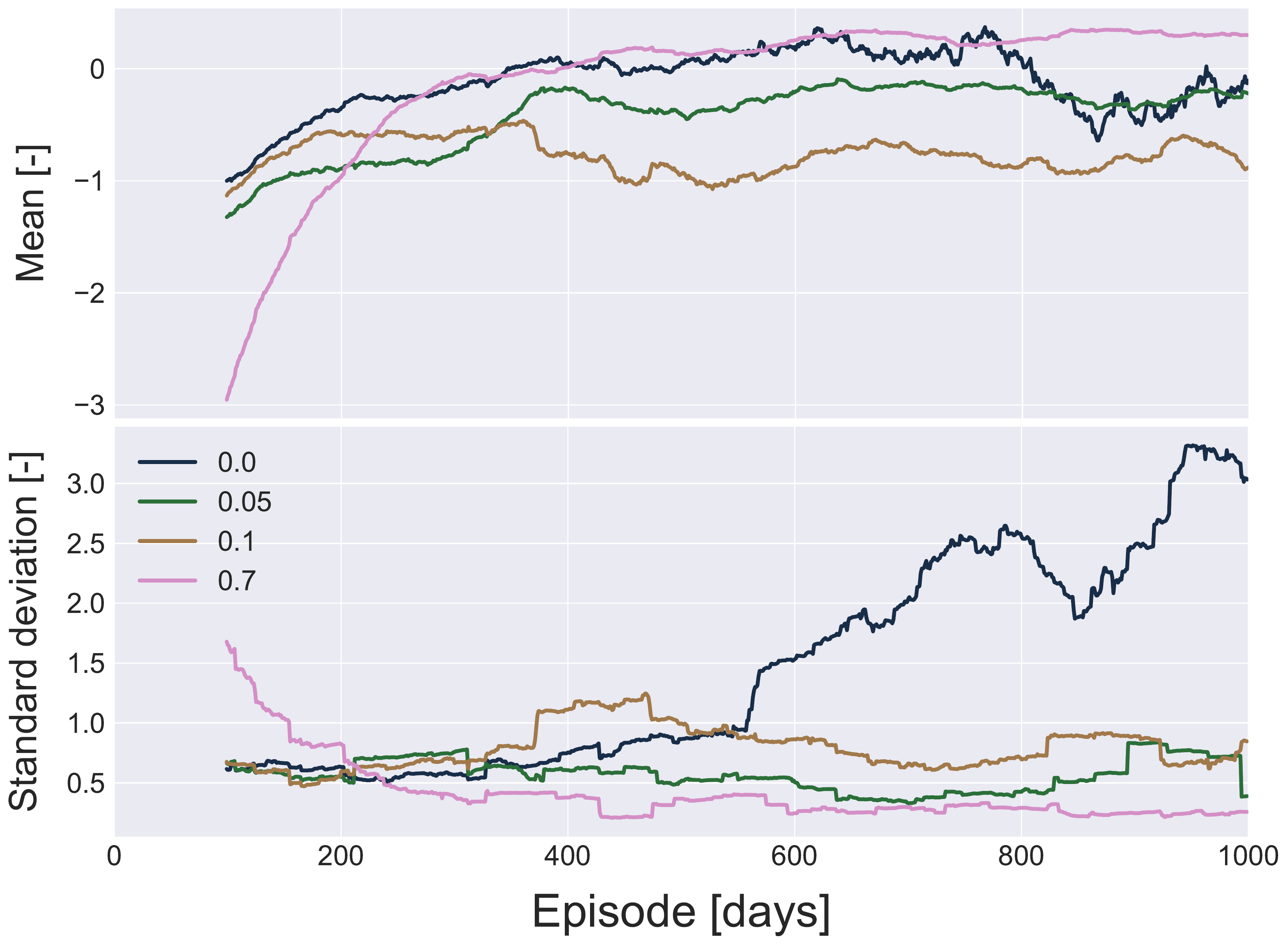}

    \caption{Rolling mean and standard deviation on the average episodic reward
    during training for increasing values of the damping factor, $\eta$,
    evaluated on \ticker{HSBA.L}.}\label{fig:eta_training}
\end{figure}

It was also observed that asymmetric dampening of the reward function lead to
improved stability during learning and better asymptotic performance compared
to the basic agent. Fig.~\ref{fig:eta_training} shows how the mean and standard
deviation of episodic reward varies with increasing values of the damping
factor, $\eta$, during training. The change due to dampening is particularly
apparent from the standard deviation which is seen to diverge for $\eta=0.0$
and decrease for $\eta \geq 0.1$. This suggests that the inventory component in
Eq.~\ref{eq:incremental_pnl} not only drives behaviour, but is also the main
source of instability, where increasing values of $\eta$ appear to yield better
and more consistent performance. It is plausible that the large scale and
variation in rewards due to changes in the agent's position dominates the
spread-capture term, and makes it very difficult to learn an accurate estimate
of the value function; one solution could be to model the value function itself
using a distribution~\cite{bellemare2017distributional}.

\emph{State representation.}
Three state representations were considered: an \emph{agent-state} (as used by
the basic agent), a \emph{full-state} (agent and market variables) and a
\emph{linear combination of tile codings}. The objective was to identify and
resolve challenges associated with increased state complexity when including
market parameters, which, as seen by the performance of the full-state variant
(Table~\ref{tab:extensions:return}), were found to have a strongly negative
impact on returns. Contrary to intuition, we did not observe any considerable
improvement in performance with increased training, instead, the agent was
regularly seen to degrade and even diverge (both in episodic reward and
TD-error).

\begin{figure}
    \centering
    \includegraphics[width=0.9\linewidth]{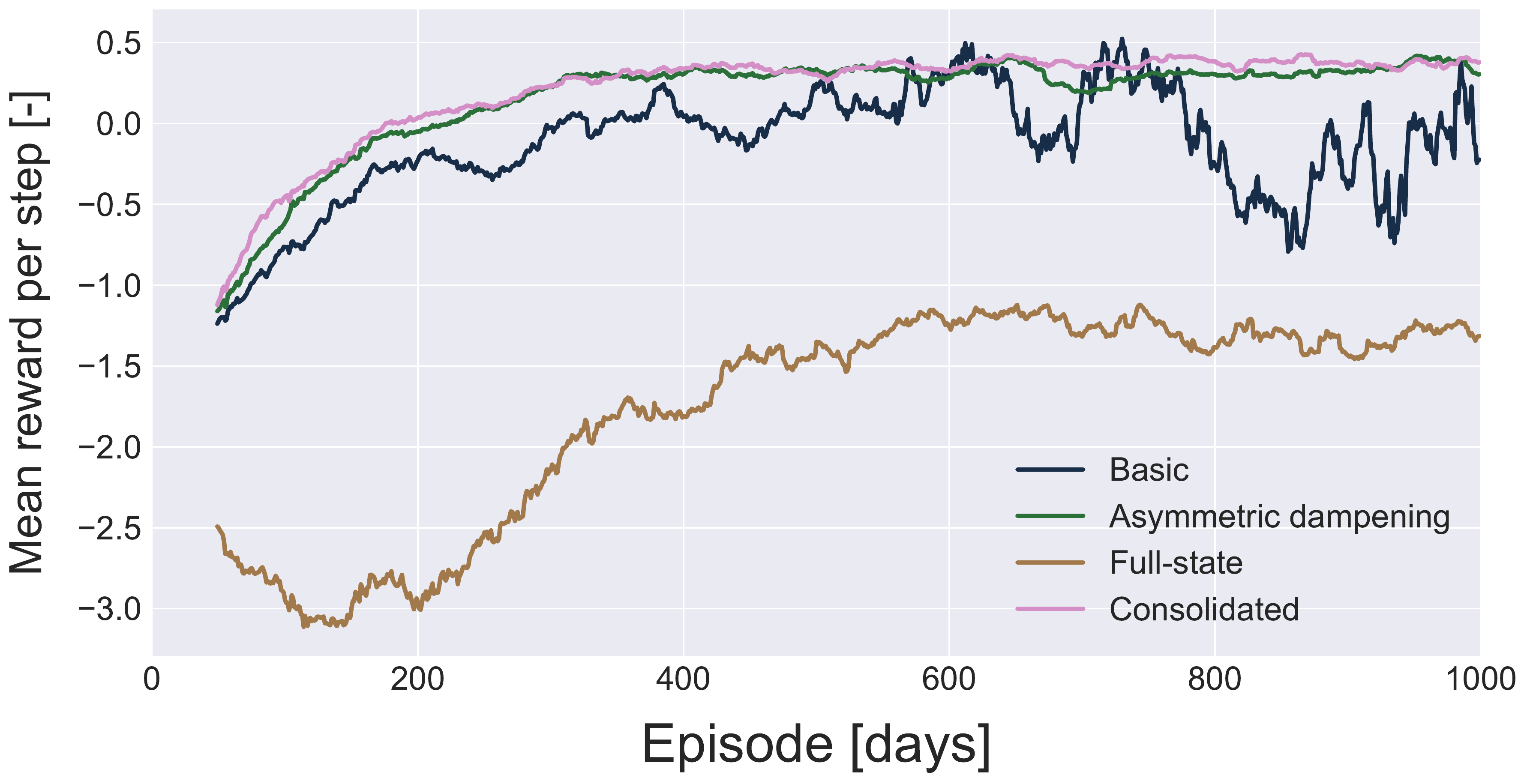}

    \caption{Rolling mean of the average episodic reward for different agent
    variants during training on \ticker{HSBA.L}.}\label{fig:training}
\end{figure}

Fig.~\ref{fig:training} shows that the basic agent is more efficient at
learning than the full-state variant and tends to be more stable; this is to be
expected since the state-space is much smaller. To capitalise on this stability
and efficiency, whilst incorporating market information, we turn to a
\emph{linear combination of tile codings (LCTC)}. As shown in
Table~\ref{tab:extensions:return}, this variant considerably outperforms the
full-state agent and doesn't appear to suffer from issues of divergence as seen
in with the full-state variant. Though in some cases it produced lower PnL than
the basic agent (as for \ticker{GSK.L}), this does not indicate a problem with
the approach, rather it is related to the choice of the variables used to
describe the state of the market; for example, we find that it performs
particularly well on \ticker{SAN.MC}. Indeed the choice of which parameters to
use for a given stock is a non-trivial task in an of itself. The key result is
that the LCTC marries expressivity and efficiency and helps to prevent
divergence even when the market variables have little informational content.

\subsection{Consolidated agent.}
\begin{table*}
    \centering
    \caption{Comparison of the out-of-sample normalised daily PnL (ND-PnL) and
    mean absolute positions (MAP) of the benchmark strategies against the final
    presented reinforcement learning agent.}\label{tab:main_comparison}

\scalebox{0.9}{%
    \setlength{\tabcolsep}{1.2em}
    \begin{tabular}{
        l
        S[table-format=-1.3,table-figures-uncertainty=1,
        separate-uncertainty=true,table-align-uncertainty=true]
        S[table-format=3,table-figures-uncertainty=1,
        separate-uncertainty=true,table-align-uncertainty=true]
        S[table-format=-1.3,table-figures-uncertainty=1,
        separate-uncertainty=true,table-align-uncertainty=true]
        S[table-format=3,table-figures-uncertainty=1,
        separate-uncertainty=true,table-align-uncertainty=true]
        S[table-format=-1.3,table-figures-uncertainty=1,
        separate-uncertainty=true,table-align-uncertainty=true]
        S[table-format=3,table-figures-uncertainty=1,
        separate-uncertainty=true,table-align-uncertainty=true]
    }
        \toprule\toprule
        & \multicolumn{2}{c}{\textbf{\citeauthor{Abernethy2013} (MMMW)}}
        & \multicolumn{2}{c}{\textbf{Fixed ($\theta_\textrm{a,b} = 5$)}}
        & \multicolumn{2}{c}{\textbf{Consolidated Agent}}
        \\
        & \multicolumn{2}{c}{\textbf{Benchmark}}
        & \multicolumn{2}{c}{\textbf{Benchmark}}
        &
        \\
        & {ND-PnL [$10^4$]} & {MAP [units]}
        & {ND-PnL [$10^4$]} & {MAP [units]}
        & {ND-PnL [$10^4$]} & {MAP [units]}
        \\
        \midrule
        \ticker{CRDI.MI} & -1.44 \pm 22.78 & 7814 \pm 1012 & -0.14 \pm 1.63 & 205 \pm 351 &
        \bfseries 0.15 \pm 0.59 & 1 \pm 2 \\
        \ticker{GASI.MI} & -1.86 \pm 9.22 & 5743 \pm 1333 & \bfseries 0.01 \pm 1.36 & 352 \pm
        523 & 0.00 \pm 1.01 & 33 \pm 65 \\
        \ticker{GSK.L} & -3.36 \pm 13.75 & 8181 \pm 1041 & 0.95 \pm 2.86 & 1342 \pm 1210 &
        \bfseries 7.32 \pm 7.23 & 57 \pm 105 \\
        \ticker{HSBA.L} & 1.66 \pm 22.48 & 7330 \pm 1059 & 2.80 \pm 10.30 & 2678 \pm 1981 &
        \bfseries 15.43 \pm 13.01 & 104 \pm 179 \\
        \ticker{ING.AS} & -6.53 \pm 41.85 & 7997 \pm 1265 & \bfseries 3.44 \pm 23.24 & 2508
        \pm 1915 & -3.21 \pm 29.05 & 10 \pm 20 \\
        \ticker{LGEN.L} & -0.03 \pm 11.42 & 5386 \pm 1297 & 0.84 \pm 2.45 & 986 \pm 949 &
        \bfseries 4.52 \pm 8.29 & 229 \pm 361 \\
        \ticker{LSE.L} & -2.54 \pm 4.50 & 4684 \pm 1507 & 0.20 \pm 0.63 & 382 \pm 553 &
        \bfseries 1.83 \pm 3.32 & 72 \pm 139 \\
        \ticker{NOK1V.HE} & -0.97 \pm 8.20 & 5991 \pm 1304 & \bfseries -0.52 \pm 4.16 & 274
        \pm 497 & -5.28 \pm 33.42 & 31 \pm 62 \\
        \ticker{SAN.MC} & -2.53 \pm 26.51 & 8865 \pm 671 & 1.52 \pm 11.64 & 3021 \pm 2194 &
        \bfseries 5.67 \pm 13.41 & 4 \pm 9 \\
        \ticker{VOD.L} & 1.80 \pm 22.83 & 7283 \pm 1579 & 1.26 \pm 4.60 & 1906 \pm 1553 &
        \bfseries 5.02 \pm 6.35 & 46 \pm 87 \\
        \bottomrule\bottomrule
    \end{tabular}%
}
\end{table*}

Finally, a consolidation of the best variants on the basic agent is considered.
It uses the asymmetrically dampened reward function with a LCTC state-space,
trained using SARSA.\ In the majority of cases it was found that this agent
generated slightly lower returns than the best individual variants, but had
improved out-of-sample stability; see Table~\ref{tab:main_comparison}. In
addition, they tended to hold smaller inventories, which may have been a
contributing factor towards the reduced uncertainty on PnL. Though the results
vary slightly across the basket of securities, the consolidated agent was found
to produce superior risk-adjusted performance over the basic agent and extended
variants overall.

\begin{figure}
    \centering
    \includegraphics[width=0.9\linewidth]{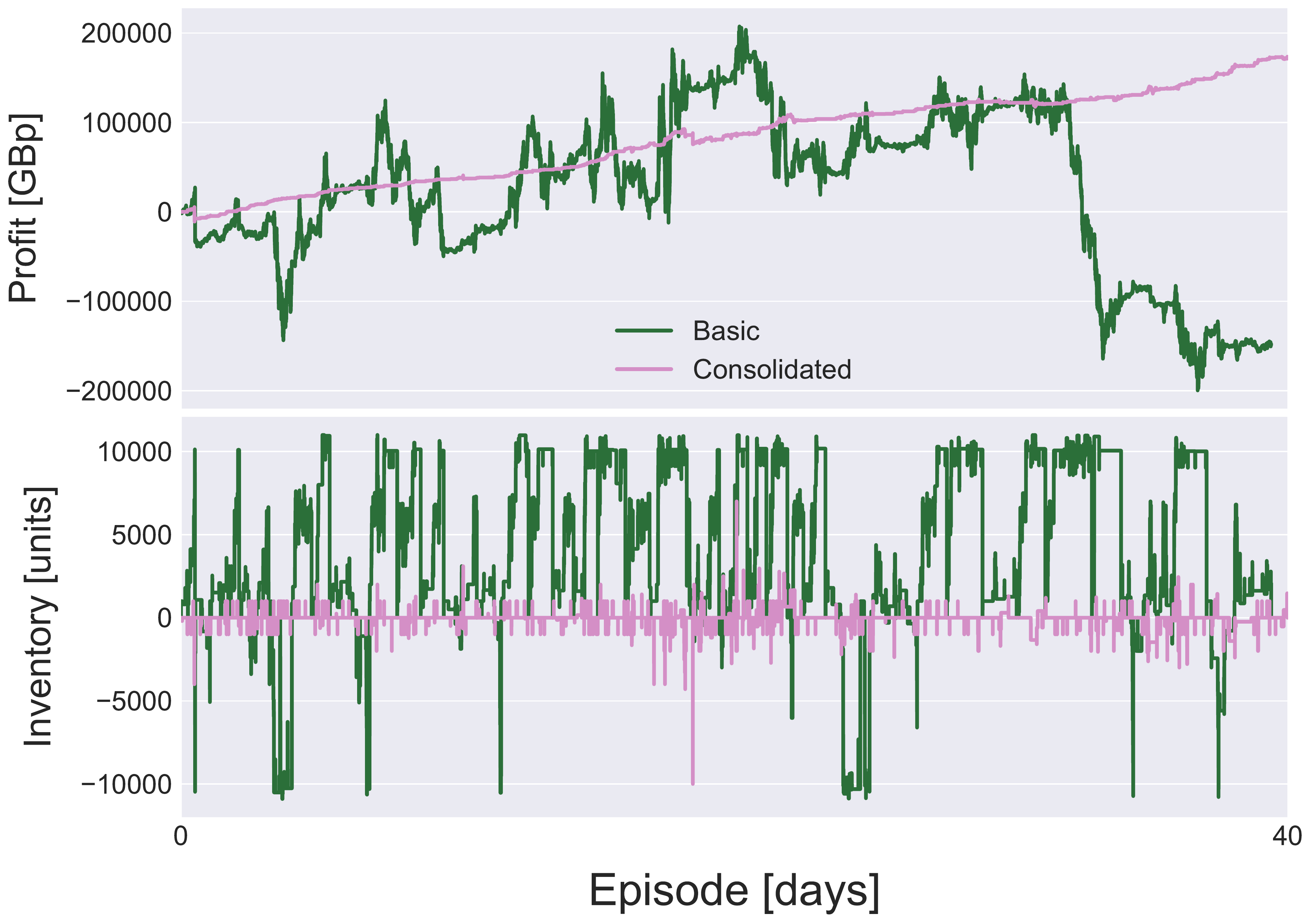}

    \caption{Out-of-sample equity curve and inventory process for the basic and
    consolidated agents, evaluated on \ticker{HSBA.L}.\vspace{3em}}\label{fig:oos}
\end{figure}

For example, Fig.~\ref{fig:oos} compares the equity and inventory processes for
the basic and consolidated agents' out-of-sample tests. Both illustrate a
profound difference in behaviour between the two. Where the former is highly
volatile, the latter is stable. The basic agent regularly holds a non-zero
inventory, exposing itself to changes in the security's value for extended
periods of time, leading to the noise observed in the equity curve. For the
consolidated agent, it appears that the learnt policy targets a near-zero
inventory, relying less on speculative trading and thus yielding the
consistency one expects from a good market making strategy.

%
\section{Conclusions}
We have developed a reinforcement learning agent that produces
competitive out-of-sample performance across a basket of securities. We first
developed a highly realistic simulation of the problem domain and showed how
eligibility traces solve the problems raised in past works. We then investigated
different learning algorithms, reward functions and state representations, and
consolidated the best techniques into a single agent which is shown to produce
superior risk-adjusted performance. The following represent key areas for 
future research based on our results:

\begin{enumerate}[leftmargin=5mm]
    \item Apply more advanced learning algorithms such as GreedyGQ, Q$(\sigma)$
        and true online variants~\cite{Maei2010,Asis2017,Seijen2016}, which
        provide guarantees of convergence with linear function approximation,
        options learning~\cite{Bacon2016} and direct policy search methods.
    \item Explore deep reinforcement learning, in particular using recurrent
        neural networks that should be well-suited to the sequential nature of
        the problem.
    \item Introduce parametrised action spaces~\cite{Masson2016} as an
        alternative to discrete action sets.
    \item Extend to multiple order and variable order size action spaces.
    \item Investigate the impact of market frictions such as rebates, fees and
        latency on the agent's strategy.
    \item Use sequential Bayesian methods~\cite{Christensen2013} for better
        order book reconstruction and estimation of order queues.
\end{enumerate}

%
\newpage
\bibliographystyle{ACM-Reference-Format}
\bibliography{main}

%
\newpage
\begin{appendices}
    \section{Appendix}
    \begin{table}[!ht]
    \centering
    \caption{Out-of-sample normalised daily PnL (ND-PnL) and mean absolute
    positions (MAP) of the FTL benchmark strategy derived
    from~\cite{Abernethy2013}.}\label{tab:ftl}

    \setlength{\tabcolsep}{1.2em}
    \begin{tabular}{
        l
        S[table-format=-1.3,table-figures-uncertainty=1,
        separate-uncertainty=true,table-align-uncertainty=true]
        S[table-format=3,table-figures-uncertainty=1,
        separate-uncertainty=true,table-align-uncertainty=true]
    }
        \toprule\toprule
        & {ND-PnL [$10^4$]} & {MAP [units]}
        \\
        \midrule
        \ticker{CRDI.MI} & -14.93 \pm 25.52 & 5705 \pm 1565 \\
        \ticker{GASI.MI} & -8.50 \pm 24.00 & 6779 \pm 1499 \\
        \ticker{GSK.L} & -29.33 \pm 95.39 & 8183 \pm 1272 \\
        \ticker{HSBA.L} & -4.00 \pm 35.84 & 8875 \pm 683 \\
        \ticker{ING.AS} & -27.53 \pm 114.97 & 9206 \pm 981 \\
        \ticker{LGEN.L} & 0.29 \pm 12.45 & 5824 \pm 1512 \\
        \ticker{LSE.L} & -2.60 \pm 4.49 & 4776 \pm 1615 \\
        \ticker{NOK1V.HE} & -3.47 \pm 8.81 & 5662 \pm 1533 \\
        \ticker{SAN.MC} & -8.80 \pm 50.00 & 9273 \pm 470 \\
        \ticker{VOD.L} & -1.72 \pm 25.11 & 8031 \pm 1610 \\
        \bottomrule\bottomrule
    \end{tabular}%
\end{table}

\end{appendices}

\end{document}